\documentclass[journal]{IEEEtran}

\usepackage{times}
\usepackage{helvet}
\usepackage{courier}

\usepackage{times}
\usepackage{xcolor}
\usepackage{soul}

\usepackage[small]{caption}
\usepackage{graphicx}
\usepackage{subfigure}
\usepackage{graphics,epsfig,amsmath,amsfonts,amssymb,times,colortbl,subfigure,floatflt,wrapfig,enumerate,tikz,url,xcolor}
\usepackage{multirow}
\usepackage{algorithmic,eqparbox,comment}
\usepackage{bm}
\usepackage{booktabs}
\usepackage[symbol]{footmisc}
\usepackage{algorithm, algorithmic}
\usepackage{array}

\newtheorem{definition}{Definition}

\newtheorem{alg}{Algorithm}

\newcommand{\mat}[1]{\bm{#1}}
\newcommand{\ten}[1]{\bm{\mathcal{#1}}}

\usepackage[utf8]{inputenc} 
\usepackage[T1]{fontenc}    
\usepackage{hyperref}       
\usepackage{url}            
\usepackage{booktabs}       
\usepackage{amsfonts}       
\usepackage{nicefrac}       
\usepackage{microtype}      

\begin{document}
\bstctlcite{IEEEexample:BSTcontrol}
\title{Matrix Product Operator Restricted Boltzmann Machines}
\author{Cong Chen, 
      Kim Batselier, 
      Ching-Yun Ko, 
      and~Ngai Wong
      \\
chencong@eee.hku.hk,
k.batselier@tudelft.nl,
cyko@eee.hku.hk,
nwong@eee.hku.hk
}
\maketitle
\begin{abstract}
A restricted Boltzmann machine (RBM) learns a probability distribution over its input samples and has numerous uses like dimensionality reduction, classification and generative modeling. Conventional RBMs accept vectorized data that dismisses potentially important structural information in the original tensor (multi-way) input. Matrix-variate and tensor-variate RBMs, named MvRBM and TvRBM, have been proposed but are all restrictive by model construction, which leads to a weak model expression power. This work presents the matrix product operator RBM (MPORBM) that utilizes a tensor network generalization of Mv/TvRBM, preserves input formats in both the visible and hidden layers, and results in higher expressive power. A novel training algorithm integrating contrastive divergence and an alternating optimization procedure is also developed. Numerical experiments compare the MPORBM with the traditional RBM and MvRBM for data classification and image completion and denoising tasks. The expressive power of the MPORBM as a function of the MPO-rank is also investigated. 
\end{abstract}
\begin{IEEEkeywords}
tensors, matrix product operators, restricted Boltzmann machines
\end{IEEEkeywords}

\section{Introduction}
\label{intro}

\IEEEPARstart{A} restricted Boltzmann machine (RBM)~\cite{hinton2002training} is a probabilistic model that employs a layer of hidden variables to achieve highly expressive marginal distributions. RBMs are an unsupervised learning technique and have been extensively explored and applied in various fields, such as pattern recognition~\cite{larochelle2008classification}, computer vision~\cite{krizhevsky2010factored} and signal processing~\cite{mohamed2010phone}. However, conventional RBMs are designed for vector data and cannot directly deal with matrices and higher-dimensional data, which are common in real-life. For example, grayscale images are second-order tensors (i.e. matrices) while color images or grayscale videos are third-order tensors. The traditional approach to apply an RBM on such high-dimensional data is through vectorization of the data which leads to two drawbacks. First, the spatial information in the original data is lost, thus weakening the model's capability to represent these structural correlations. Second, the fully connected visible and hidden units may cause overfitting since the intrinsic low-rank property of many real-life data is disregarded.

To address these, we propose a matrix product operator (MPO) restricted Boltzmann machine (MPORBM) where both the visible and hidden layers are in tensor forms. The link between the visible and hidden layers is represented by an MPO, which is essentially a tensor network representation of a matrix. To train an MPORBM, we further describe its customized parameter learning algorithms. The MPORBM is also compared with the standard RBM and matrix-variate RBM~\cite{qi2016matrix} for tensor inputs in numerical experiments.
This article has the following major contributions:
\begin{enumerate}
\item The MPORBM is proposed for the first time and generalizes existing RBM architectures. The standard RBM, matrix-variate RBM (MvRBM)~\cite{qi2016matrix} and tensor-variate RBM (TvRBM)~\cite{nguyen2015tensor} models are all special cases of the MPORBM.
\item Compared with standard RBMs, the number of parameters in an MPORBM grows only linearly with the order of the tensor instead of exponentially. This alleviates overfitting, which is demonstrated through numerical experiments.
\item Both the visible and hidden layers are represented in tensor forms and therefore useful structural information in the original tensor data is preserved and utilized.
\item The graphical ``language" of tensor network diagrams~\cite{orus2014practical} is introduced to represent how specific quantities (such as the conditional probabilities) are computed. This way, complicated mathematical expressions are easily understood in a visual manner.
\item Although the data structures are generalized to tensors, the bipartite graph nature of an RBM still applies, together with the fast sampling and inference properties. 
\end{enumerate}


\section{Related work}
\label{relatedWork}
Real-life data are extensively multiway. Researchers have been motivated to develop corresponding multiway RBMs. For example,~\cite{taylor2009factored} proposed a factored conditional RBM for modeling motion style. In their model, both historical and current motion vectors are considered as inputs so that the pairwise association between them is captured. However, since the visible layer is in vector form, the spatial information in the multiway data is not retained. In~\cite{krizhevsky2010factored} a three-way factored conditional RBM was proposed where a three-way weight tensor is employed to capture the correlations between the input, output, and hidden variables. However, their training data still requires vectorization. 

The above works are both aiming to capture the interaction among different vector inputs and are hence not directly applicable to matrix and tensor data. The first RBM designed for tensor inputs is~\cite{nguyen2015tensor}, which is called a tensor-variate RBM (TvRBM). In TvRBM, the visible layer is represented as a tensor but the hidden layer is still a vector. Furthermore, the connection between the visible and hidden layers is described by a canonical polyadic (CP) tensor decomposition~\cite{kolda2009tensor}. However, this CP weight tensor is claimed to constrain the model representation capability~\cite{qi2016matrix}.

Another RBM related model that utilizes tensor input is the matrix-variate RBM (MvRBM)~\cite{qi2016matrix}. The visible and hidden layers in an MvRBM are both matrices. Nonetheless, to limit the number of parameters, an MvRBM models the connection between the visible and hidden layers through two separate matrices, which restricts the ability of the model to capture correlations between different data modes.

All these issues have motivated the MPORBM. Specifically, MPORBM not only employs tensorial visible and hidden layers, but also utilizes a general and powerful tensor network, namely an MPO, to connect the tensorial visible and hidden layers. By doing so, an MPORBM achieves a more powerful model representation capacity than MvRBM and at the same time greatly reduces the model parameters compared to a standard RBM. Note that a mapping of the standard RBM with tensor networks has been described in~\cite{Chen2018}. However, their work does not generalize the standard RBM to tensorial inputs and is therefore still based on visible and hidden units in vector forms.

\section{Preliminaries}
\label{preliminary}
We review some necessary tensor basics, the MPO tensor decomposition, the standard RBM and its tensorial variants. 
\subsection{Tensor basics}
\label{tensor}
\begin{figure}[t] 
\begin{center} 
\includegraphics[width=0.45\textwidth]{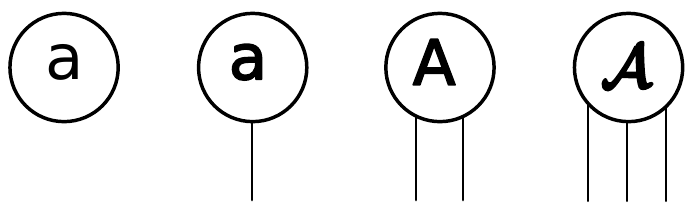}
\caption{Graphical representation of a scalar $a$, vector $\mat{a}$, matrix $\mat{A}$, and third-order tensor $\ten{A}$.}
\label{fig:graphical}
\end{center}
\end{figure}

Tensors are multi-dimensional arrays that are higher-order generalization of vectors (first-order tensors) and matrices (second-order tensors). A $d$th-order (also called $d$-way or $d$-mode) tensor is denoted as $\ten{A}\in\mathbb{R}^{I_1\times I_2 \times \cdots \times I_d}$ where each entry $\ten{A}(i_1,i_2,\ldots,i_d)$ is determined by $d$ indices \mbox{$1 \leq i_k \leq I_k (k = 1,2,\ldots,d)$}. The numbers $I_1, I_2,\ldots, I_d$ are called the dimensions of the tensor.
We use boldface capital calligraphic letters $\ten{A}$, $\ten{B}$, \ldots to denote tensors, boldface capital letters $\mat{A}$, $\mat{B}$, \ldots to denote matrices, boldface letters $\mat{a}$, $\mat{b}$, \ldots to denote vectors, and roman letters $a$, $b$, \ldots to denote scalars. $\mat{A}^T$ and $\mat{a}^T$ are the transposes of a matrix $\mat{A}$ and a vector $\mat{a}$. An intuitive and useful graphical representation, named tensor network diagrams, of scalars, vectors, matrices and tensors is depicted in Fig.~\ref{fig:graphical}. The unconnected edges are the indices of the tensor. We will mainly employ these graphical representations to visualize the tensor networks and operations in the following sections and refer to~\cite{orus2014practical} for more details. We now briefly introduce some important tensor operations.

\begin{figure}[ht] 
\begin{center} 
\includegraphics[width=0.4\textwidth]{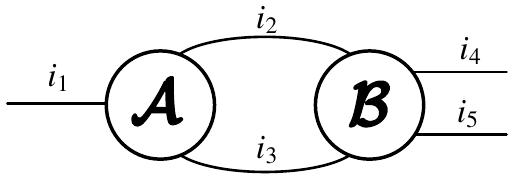}
\caption{Tensor contraction between a third-order tensor $\ten{A}$ and a fourth-order tensor $\ten{B}$, where the summation runs over  $i_2$ and $i_3$.}
\label{fig:indexcontraction}
\end{center}
\end{figure}

\begin{definition}(Tensor index contraction): A tensor index contraction is the sum over all possible values of the repeated indices in a set of tensors. 
\end{definition}
For example, the following contraction between a third-order tensor $\ten{A}$ and a fourth-order tensor $\ten{B}$
\begin{align}
 \ten{C}(i_1,i_4,i_5)&=
\nonumber  \sum\limits_{i_2=1}^{I_2}\sum\limits_{i_3=1}^{I_3}  \ten{A}(i_1,i_2,i_3)\, \ten{B}(i_2,i_3,i_4,i_5),
\end{align}
over the $i_2$ and $i_3$ indices results in a third-order tensor $\ten{C}$. The corresponding tensor network diagram for these index contractions is shown in  Fig.~\ref{fig:indexcontraction}, where the summation over the $i_2$ and $i_3$ indices is indicated by the connected edges. The resulting diagram has three unconnected indices ($i_1,i_4,i_5$), which confirms that $\ten{C}$ is of third order.


\subsection{Tensor MPO decomposition}
\label{MPOdecom}
An MPO is essentially a tensor network representation of a matrix. To relate the row and column indices of a matrix to the corresponding multi-indices of the MPO we need the following definition.
\begin{definition}
The mapping between a linear index $i$ of a vector $\mat{a} \in \mathbb{R}^{I_1\cdots I_d}$ and its corresponding multi-index $[i_1,i_2,\ldots,i_d]$ when reshaped into a tensor $\ten{A}\in \mathbb{R}^{I_1\times \cdots \times I_d}$ is %
\begin{align}
i &= 
i_1+\sum_{k=2}^d{(i_{k}-1)\prod_{p=1}^{k-1}I_{p}}.
\label{eqn:single2multi}
\end{align}
\end{definition}

\begin{figure*}[t]
\begin{center} 
\includegraphics[width=0.8\textwidth]{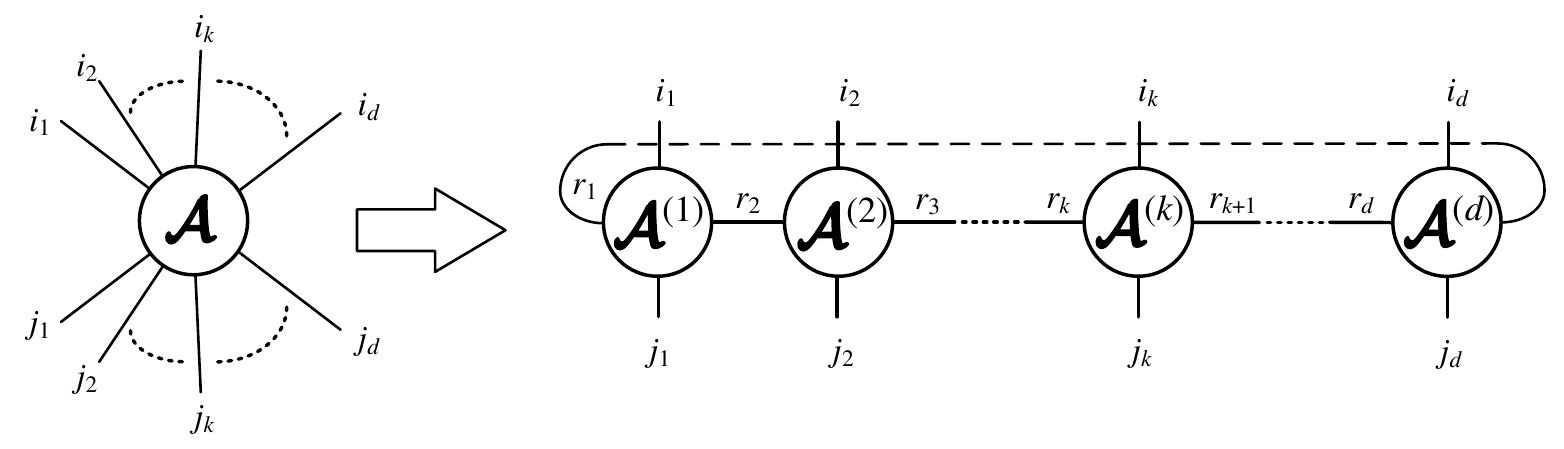}
\caption{Matrix product operator (MPO) decomposition.}
\label{fig:mpodecom}
\end{center}
\end{figure*}

Now suppose we have a matrix \mbox{$\mat{A}\in\mathbb{R}^{I_1\cdots I_d\times J_1\cdots  J_d}$}, where the index mapping~\eqref{eqn:single2multi} is used to split both the row index $i$ and column index $j$ into multi-indices $[i_1,\ldots,i_d],[j_1,\ldots,j_d]$, respectively. After this index splitting we can arrange all matrix entries into a $2d$-way tensor $\ten{A}\in\mathbb{R}^{I_1\times\cdots \times I_d\times J_1\times\cdots \times J_d}$. Per definition, the corresponding MPO decomposition represents each entry of $\ten{A}$ as 
\begin{equation}
\begin{split}
\label{eqn:mpo}
&\ten{A}(i_1,\ldots,i_d,j_1,\ldots,j_d)=  \\ &\sum\limits_{r_1,r_2,\ldots,r_d}^{R_1,R_2,\ldots,R_d} \ten{A}^{(1)}(r_1,i_1,j_1,r_2) \cdots \ten{A}^{(d)}(r_{d},i_d,j_d,r_1).
\end{split}
\end{equation}
The ``building blocks" of the MPO are the $4$-way tensors $\ten{A}^{(1)},\ldots,\ten{A}^{(d)}$, also called the MPO-cores. The dimensions $R_1,R_2,\ldots,R_d$ of the summation indices $r_1, \ldots, r_d$ are called the MPO-ranks. Observe that the second summation index of $\ten{A}^{(d)}$ is the same as the first index of $\ten{A}^{(1)}$, which ensures that the right-hand side of~\eqref{eqn:mpo} results in a scalar. When $R_1 > 1$, the MPO is said to have periodic boundary conditions. We will assume throughout the remainder of this article that $R_1=1$. The tensor network diagram representation of \eqref{eqn:mpo} is shown in Fig.~\ref{fig:mpodecom}. All index contractions are again represented as edges connecting two tensors. The main motivation to use an MPO representation of a matrix is to reduce its storage requirement. Indeed, using an MPO-representation with maximal MPO-rank $R$ for a matrix $\mat{A}\in \mathbb{R}^{I^d \times I^d}$ reduces the storage requirement from $I^{2d}$ down to approximately $dI^2R^2$. This leads to significant storage savings when $R$ is relatively small. Consequently, all computations are done on the MPO-cores rather than on the whole matrix itself. This could potentially reduce the computational complexity of the learning algorithm, especially if the visible and hidden layers are also represented as matrix product states (MPS, also called tensor trains in the scientific computing community~\cite{ivanTT}). This extension is kept for future work.

\subsection{Standard RBM}
\label{RBM}
The standard RBM~\cite{hinton2002training} is a bipartite undirected probabilistic graphical model with one visible layer $\mat{v}\in \mathbb{R}^M$ and one hidden layer $\mat{h}\in \mathbb{R}^N$, both in vector forms. Here we mainly consider binary RBMs, which implies that entries in both $\mat{v}$ and $\mat{h}$ attain binary values. The standard RBM assigns the following energy function for a specific joint configuration $\{ \mat{v},\mat{h}\}$:
\begin{align}
E(\mat{v},\mat{h};\Theta)=-\mat{v}^T\mat{W}\mat{h}-\mat{v}^T\mat{b}-\mat{c}^T\mat{h},
\label{eq:RBMenergyfun}
\end{align}
where $\mat{b}\in \mathbb{R}^M $ and $\mat{c}\in \mathbb{R}^N $ are the bias of the visible layer and hidden layer, respectively, and $\mat{W}\in \mathbb{R}^{M \times N} $ is the mapping matrix. All model parameters together are denoted $\Theta = \{ \mat{W},\mat{b},\mat{c}\}$. We can easily use a tensor network diagram to represent Eq.~\ref{eq:RBMenergyfun} in Fig.~\ref{fig:energyProb} (a). The conditional distributions over the hidden and visible layers can be written as:
\begin{align}
p(\mat{v}=\mat{1}|\mat{h};\Theta)=\sigma(\mat{W}\mat{h}+\mat{b}),
\label{eq:RBMh2v}
\end{align}
\begin{align}
p(\mat{h}=\mat{1}|\mat{v};\Theta)=\sigma(\mat{W}^T\mat{v}+\mat{c}),
\label{eq:RBMv2h}
\end{align}
where $\sigma(x)=1/(1+e^{-x})$ is the logistic sigmoid function and $\mat{1}$ denotes a vector of ones. The parameter training in a standard RBM is commonly performed by the contrastive divergence (CD) algorithm~\cite{hinton2002training} and its variants~\cite{tieleman2008training,tieleman2009using}.  

\subsection{Matrix-variate and tensor-variate RBMs}

Here we briefly introduce the two non-vector RBM models, namely MvRBM~\cite{qi2016matrix} and TvRBM~\cite{nguyen2015tensor}. TvRBM employs a tensorial visible layer and keeps the hidden layer in a vector form. A rank-$R$ CP tensor decomposition is used to connect the visible and hidden layers. However, such a connection form is also criticized to limit the model capability~\cite{qi2016matrix}. The corresponding energy function of TvRBM is shown in Fig.~\ref{fig:energyProb}(b) as a tensor network diagram.

In the MvRBM model, both the input and hidden layers are matrices and are interconnected through two independent weight matrices $\mat{W}^{(1)} \in \mathbb{R}^{M_1\times N_1},\mat{W}^{(2)}\in \mathbb{R}^{M_2\times N_2}$. This particular construction reduces the total number of parameters from $M_1\times M_2\times N_1\times N_2$ down to $M_1\times N_1+M_2\times N_2$ but comes at the cost of a limited representation ability, as the weight matrix $\mat{W}$ is constrained to be a Kronecker product of the $\mat{W}^{(1)},\mat{W}^{(2)}$ matrices. The energy function of the MvRBM is graphically represented as a tensor diagram in Fig.~\ref{fig:energyProb}(c).

\section{MPORBM}
\label{MPORBM}
We now describe the proposed MPORBM and its customized model parameter learning algorithms.

\subsection{Model definition}
\label{modeldef}
\begin{figure*}[t] 
\begin{center} 
\includegraphics[width=0.8\textwidth]{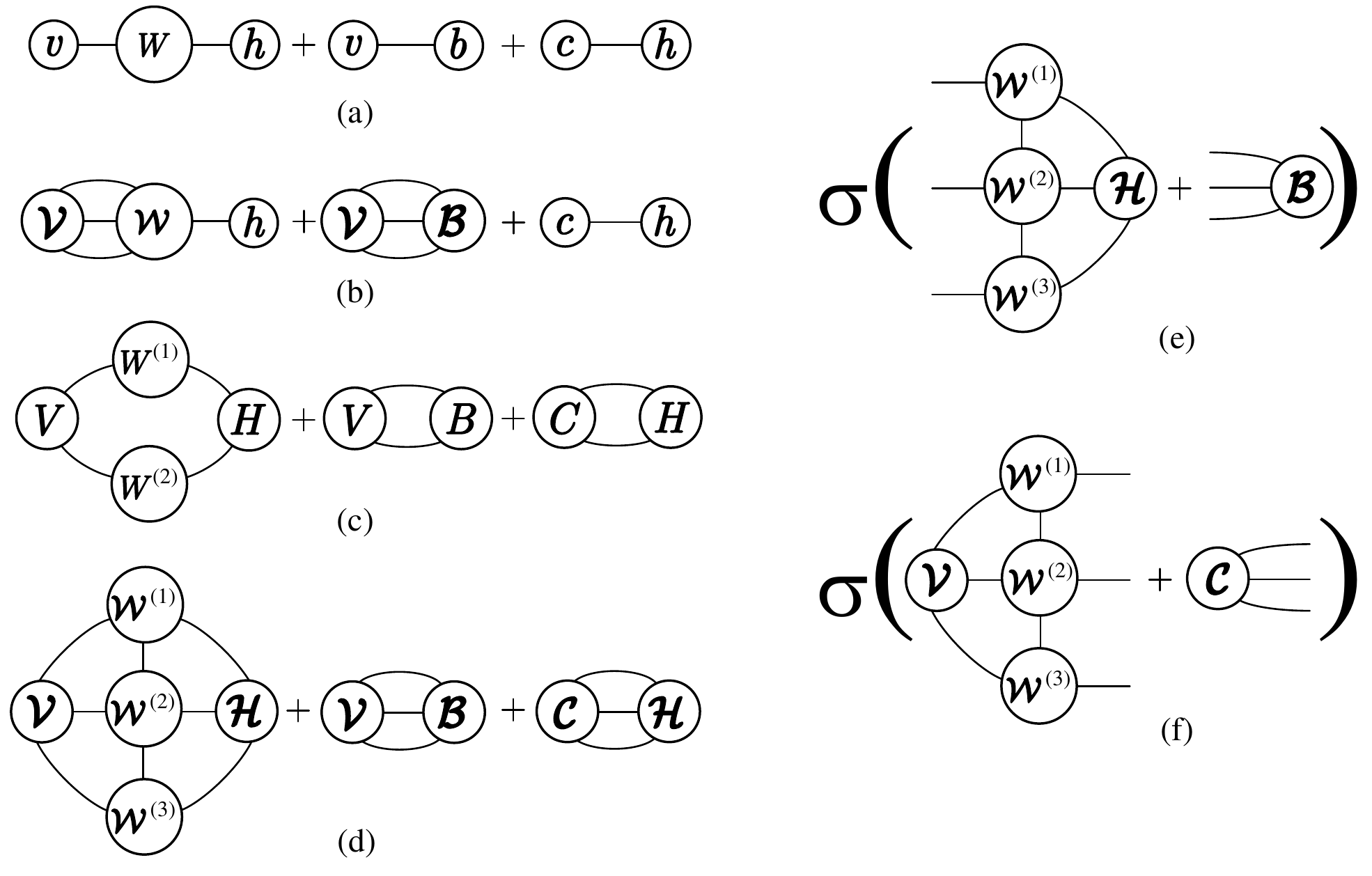}
\caption{Negative energy functions ($-E$) of (a) RBM; (b) TvRBM; (c) MvRBM; (d) MPORBM. And conditional probability of an MPORBM over its (e) hidden layer; (f) visible layer.}
\label{fig:energyProb}
\end{center}
\end{figure*}

In an MPORBM, both the visible layer \mbox{$\ten{V}\in \mathbb{R}^{I_1\times\cdots \times I_d}$} and the hidden layer \mbox{$\ten{H}\in \mathbb{R}^{J_1\times\cdots \times J_d}$} are $d$-way tensors. As a result, the weight matrix $\mat{W}$ is now a $2d$-way tensor \mbox{$\ten{W} \in \mathbb{R}^{I_1\times\cdots \times I_d\times J_1\times\cdots \times J_d}$}, which is represented by an MPO instead. With both the visible and hidden layers being tensors, it is therefore also required that the bias vectors $\mat{b},\mat{c}$ are tensors $\ten{B}\in \mathbb{R}^{I_1\times\cdots \times I_d}, \ten{C}\in \mathbb{R}^{J_1\times\cdots \times J_d}$, respectively. The energy function of an MPORBM is shown in Fig.~\ref{fig:energyProb}(d) for the specific case where $d=3$. The vertical edges between the different MPO-cores $\ten{W}^{(1)},\ldots, \ten{W}^{(d)}$ are the key ingredient in being able to express generic weight tensors $\ten{W}$, as opposed to the two disconnected weight matrices in the MvRBM model. The corresponding conditional distribution equations over the hidden and visible layers can be derived from the energy function and are visualized in Fig.~\ref{fig:energyProb}(e)\&(f) for the case where $d=3$. This involves the summation of the weight MPO with either the hidden or visible layer tensors into a $d$-way tensor, which is then added elementwise with the corresponding bias tensor. The final step in the computation of the conditional probability is an elementwise application of the logistic sigmoid function on the resulting tensor.

\subsection{MPORBM Learning algorithm}
\label{learningalgo}

Let $\Theta=\{\ten{B}, \ten{C}, \ten{W}^{(1)}, \ten{W}^{(2)}, \ldots, \ten{W}^{(d)}\}$ denote the model parameters. The model learning task is then formulated into maximizing the training data likelihood: 
\begin{align}
\ten{L}(\ten{V};\Theta)= p(\ten{V};\Theta)=  \sum\limits_{\ten{H}}p(\ten{V},\ten{H};\Theta)
\label{eq:likelihood}
\end{align}
with respect to model parameter $\Theta$. Similar to the standard RBM~\cite{hinton2002training}, the expression of the gradient of the log-likelihood is:
\begin{align}
\small
\frac{\partial}{\partial \Theta}\textrm{log}\ten{L}(\ten{V};\Theta) = - \mathbb{E}_{\ten{H}|\ten{V}}\bigg[ \frac{\partial {E(\ten{V},\ten{H})}}{\partial \Theta} \bigg] + \mathbb{E}_{\ten{V},\ten{H}}\bigg[ \frac{\partial {E(\ten{V},\ten{H})}}{\partial \Theta} \bigg]
\label{eq:loglikelihood}
\end{align}
where $\mathbb{E}_{\ten{V}|\ten{H}}$ is the data expectation w.r.t. $p(\ten{H}|\ten{V};\Theta)$, which can be computed by Fig.~\ref{fig:energyProb}(f). $\mathbb{E}_{\ten{V},\ten{H}}$ is the model expectation w.r.t. $p(\ten{H},\ten{V};\Theta)$, which can be approximately computed by the CD procedure. The main idea in the CD procedure is as follows: first, a Gibbs chain is initialized with one particular training sample $\ten{V}_{(0)}$= $\ten{X}_{train}$. Figs.~\ref{fig:energyProb}(e)\&(f) are then computed $K$ times in an alternating fashion, which results in the chain $\{ (\ten{V}_{(0)},\ten{H}_{(0)}), (\ten{V}_{(1)},\ten{H}_{(1)}),\ldots, (\ten{V}_{(K)},\ten{H}_{(K)})\}$. The model expectation is then approximated by $\{ \ten{V}_{(K)}\}$. 
\begin{figure*}
\begin{center} 
\includegraphics[width=0.8\textwidth]{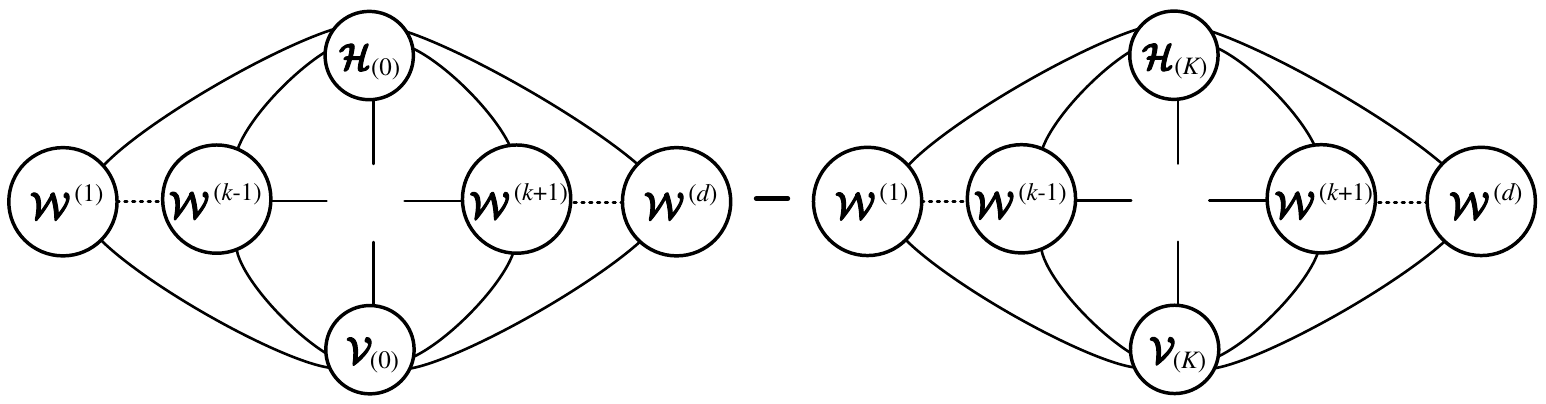}
\caption{The derivatives of the log-likelihood function with respect to the $k$th MPO-core $\ten{W}^{(k)}$.}
\label{fig:deriv_Wi}
\end{center}
\end{figure*}
The derivative of the log-likelihood with respect to the $k$-th MPO-core $\ten{W}^{(k)}$ is visualized in Fig.~\ref{fig:deriv_Wi}. This involves the computation of 2 fourth-order tensors, obtained by removing the $k$-th MPO-core $\ten{W}^{(k)}$ from the two tensor networks and summing over all connected edges. The resulting tensors are then subtracted elementwise from one another. The derivatives of the log-likelihood with respect to the bias tensors $\ten{B},\ten{C}$ are
\begin{align*}
\frac{\partial}{\partial \ten{B}}\textrm{log}\ten{L}(\ten{V};\Theta) = \ten{V}_{(0)}- \ten{V}_{(K)},
\end{align*}

\begin{align*}
\frac{\partial}{\partial \ten{C}}\textrm{log}\ten{L}(\ten{V};\Theta) = \ten{H}_{(0)}- \ten{H}_{(K)}.
\end{align*}

We mainly use the CD procedure to train the MPORBM model. However, instead of updating all the MPO-cores simultaneously with one batch of input training data, we employ the alternating optimization procedure (AOP). This involves updating only one MPO-core at a time while keeping the others unchanged using the same batch of input training data. We name this parameter learning algorithm CD-AOP and its pseudo-code is given in Algorithm~\ref{alg:MPORBM} . The superiority of this alternating optimization procedure over simultaneously updating all MPO-cores, which we call CD-SU from hereon will be demonstrated through numerical experiments.
\begin{alg}MPORBM learning algorithm (CD-AOP)\\
\label{alg:MPORBM}
\textit{\textbf{Input}}: Training data of N tensors $\ten{D}=\{ \ten{X}_1, \ldots, \ten{X}_N\}$, the maximum iteration number $T$, the batch size $b$, the momentum $\gamma$, the learning rate $\alpha$ and CD step $K$.\\
\textit{\textbf{Output}}: Model parameters $\Theta=\{\ten{B}, \ten{C}, \ten{W}^{(1)}, \ten{W}^{(2)}, \ldots, \ten{W}^{(d)}\}$.

\begin{algorithmic}[1]
\STATE Randomly initialize $\ten{W}^{(1)},\ldots,\ten{W}^{(d)}$. Set the bias $\ten{B}=\ten{C}=\ten{O}$ and the gradient increments $\Delta \ten{B} = \Delta \ten{C}= \Delta \ten{W}^{(1)}= \cdots =\Delta \ten{W}^{(d)}=\ten{O}$. 
\FOR {iteration number $t=1 \rightarrow T$}
\STATE Randomly divide $\ten{D}$ into M batches $\ten{D}_1, \ldots, \ten{D}_M$ of size $b$.
\FOR {batch $m=1\rightarrow M$}
\FOR { $c=1 \rightarrow d$} 
\STATE For all data $\ten{V}_{(0)}=\ten{X}_{train} \in \ten{D}_m $ run Gibbs sampling:
\FOR {k=0,\ldots,K-1}
\STATE sample $\ten{H}_{(k)}$ according to Fig.~\ref{fig:energyProb} (f) with $\ten{V}_{(k)}$;
\STATE sample $\ten{V}_{(k+1)}$ according to Fig.~\ref{fig:energyProb}(e) with $\ten{H}_{(k)}$
\ENDFOR
\STATE \small $\Delta \ten{W}^{(c)} \gets \gamma \Delta\ten{W}^{(c)}+ \alpha (\frac{1}{b} \sum\limits_{\ten{D}_m} \frac{\partial}{\partial \ten{W}^{(c)}}\textrm{log}\ten{L}(\ten{V};\Theta))$
\STATE $\Delta \ten{B} \gets \gamma \Delta\ten{B}+ \alpha (\frac{1}{b} \sum\limits_{\ten{D}_m} \frac{\partial}{\partial \ten{B}}\textrm{log}\ten{L}(\ten{V};\Theta))$
\STATE $\Delta \ten{C} \gets \gamma \Delta\ten{C}+ \alpha (\frac{1}{b} \sum\limits_{\ten{D}_m} \frac{\partial}{\partial \ten{C}}\textrm{log}\ten{L}(\ten{V};\Theta))$ 
\STATE $\Theta  \gets \Theta+\Delta \Theta$
\ENDFOR
\ENDFOR
\ENDFOR
\end{algorithmic}
\end{alg}

\section{Experiments}
\label{experiments}


In this section, the MPORBM is compared with both the standard RBM and MvRBM. 
Specifically, we first investigate the performance and scalability of these models as a means to do feature extraction for classification, together with the influence of the MPO-ranks on the expressive power of the MPORBM. Furthermore, we demonstrate the generative capacity of MPORBM by implementing an image completion task and an image denoising task. For hyperparameter setting, we select mostly the same default values as in MvRBM paper, namely momentum $\gamma$ = $0.5$, CD step $K$ = $1$. For learning rate, we provide a set of possible value and choose the one which achieves the best validation accuracy. Moreover, we choose the same maximum iteration number and batch size for all methods.  All experiments are run in MATLAB on an Intel i5 3.2GHz desktop with 16GB RAM.







\subsection{Data Classification}
In the first experiment, we demonstrate the data classification accuracy superiority of MPORBM on extensive standard machine learning datasets, namely the Binary~Alphadigits dataset\footnote{https://cs.nyu.edu/~roweis/data.html}, normalized DrivFace\footnote{https://archive.ics.uci.edu/ml/datasets/DrivFace}, Arcene\footnote{https://archive.ics.uci.edu/ml/datasets/Arcene} and COIL-100 dataset\footnote{http://www1.cs.columbia.edu/CAVE/software/softlib/coil-100.php}. The size of those datasets vary from 320 to 49152. Since we assume binary input in our RBM setting, thus for non-binary datasets, we first use a multi-bit vector to represent each value in the original data. The additional multi-bit dimension is combined with the RGB channel if the dataset is color image. The following table~\ref{tb:datasetInfor} shows the detailed information of these datasets, while table~\ref{tb:inputStruc} and table~\ref{tb:outputStruc} show the visual and hidden layer structure of different RBM models for different datasets.

\begin{table}[h]
\newcommand{\tabincell}[2]{\begin{tabular}{@{}#1@{}}#2\end{tabular}}
\centering
\caption{\label{tb:datasetInfor}Detailed information for different datasets.}
\vspace{1ex}
\begin{tabular}{@{}lrr@{}}
Datasets & Original data size & Data value range \\
\midrule
Alphadigits & $20$ x $16$&  Binary  \\
DrivFace& $80$ x $80$ &  $0$-$255$ integer  \\
 Arcene & $10000$ x $1$ &  $0$-$924$  integer  \\
  COIL-100 & $128$ x $128$ x $3$ &  $0$-$255$ integer  \\
 \end{tabular}
 \label{tb:datasetInfor}
\end{table}

 \begin{table}[h]
\newcommand{\tabincell}[2]{\begin{tabular}{@{}#1@{}}#2\end{tabular}}
\centering
\caption{\label{tb:inputStruc}Visual layer structure of different RBM models for different datasets.}
\vspace{1ex}
\begin{tabular}{@{}lccc@{}}
Datasets & RBM & MvRBM & MPORBM  \\
\midrule
Alphadigits & $320$ x $1$&  $20$ x $16$& $20$ x $16$ \\
DrivFace & $51200$ x $1$ &  $80$ x $640$ & $80$ x $80$ x $8$ \\
Arcene & $100000$ x $1$ &  $100$ x $1000$ & $100$ x $100$ x $10$ \\
COIL-100 &  $393216$ x $1$ &  $128$ x $3072$ & $128$ x $128$ x $24$ \\
 \end{tabular}
 \label{tb:inputStruc}
\end{table}
 
  \begin{table}[H]
\newcommand{\tabincell}[2]{\begin{tabular}{@{}#1@{}}#2\end{tabular}}
\centering
\caption{\label{tb:outputStruc}Hidden layer structure of different RBM models for different datasets.}
\vspace{1ex}
\begin{tabular}{@{}lccc@{}}
Datasets  & RBM & MvRBM & MPORBM \\
\midrule
Alphadigits & $80$ x $1$&  $10$ x $8$& $10$ x $8$ \\
DrivFace & $500$ x $1$ &  $10$ x $50$ & $10$ x $10$ x $5$ \\
Arcene & $500$ x $1$ &  $10$ x $50$ & $10$ x $10$ x $5$ \\
COIL-100 & $500$ x $1$ &  $10$ x $50$ & $10$ x $10$ x $5$ \\
 \end{tabular}
 \label{tb:outputStruc}
\end{table}



We randomly separate the above four datasets into training, validation and testing parts respectively. It might happen that training sample sizes are insufficiently large to train a good RBM model. Thus the training data number we chose for each class is significant smaller than their data dimension, commonly less than 35. The MPORBM-ranks were chosen empirically from a range of 10 to 50 depended on the model input size. The trained RBM models were then employed to extract features from the hidden layer and then those new features will be utilized to train a $K$ Nearest Neighbor ($K$-NN) classifier with $K=1$ for all experiments. Table~\ref{tb:addEx1} lists the resulting classification errors.

\begin{table}[h]
\footnotesize
\newcommand{\tabincell}[2]{\begin{tabular}{@{}#1@{}}#2\end{tabular}}
\centering
\caption{\label{tb:experiment1}Classification errors of different RBM models.}
\vspace{1ex}
\begin{tabular}{@{}lccc@{}}
Datasets  & RBM & MvRBM & MPORBM CD-SU/AOP \\
\midrule
Alphadigits & $ 28.10\%$ &  $31.20\% $ & $28.10\% $ / $ 26.90\%$ \\
DrivFace & $ 24.20\% $ &  $15.48\% $ & $9.68\% $ / $ 8.06\%$ \\
Arcene & $ 45.00\% $ &  $34.00\% $ & $32.00\% $ / $ 27.00\%$ \\
COIL-100 & $ - $ &  $ 6.82\%$ & $ 6.82\% $ / $ 0.00\% $ \\
 \end{tabular}
 \label{tb:addEx1}
\end{table}

The restrictive Kronecker product assumption of the weight matrix in MvRBM explains why it has the worst classification performance in all datasets. The standard RBM also performs worse than MPORBM, which may be caused by overfitting because of the small training sample size. It is notable that due to the PC memory constraint, the standard RBM fails to learn the enormous number of parameters in the full weight matrix for COIL-100 dataset, but MPORBM still gets the best classification performance due to the general MPO weight structure. Moreover, the CD-AOP algorithm shows a higher classification accuracy than CD-SU, which further indicates that the alternating updating scheme is more suitable for our MPORBM model.

\subsection{Investigation of influence MPO-rank on classification error}
\label{digitclassification}

\begin{figure}[h]
\begin{center} 
\includegraphics[width=0.5\textwidth]{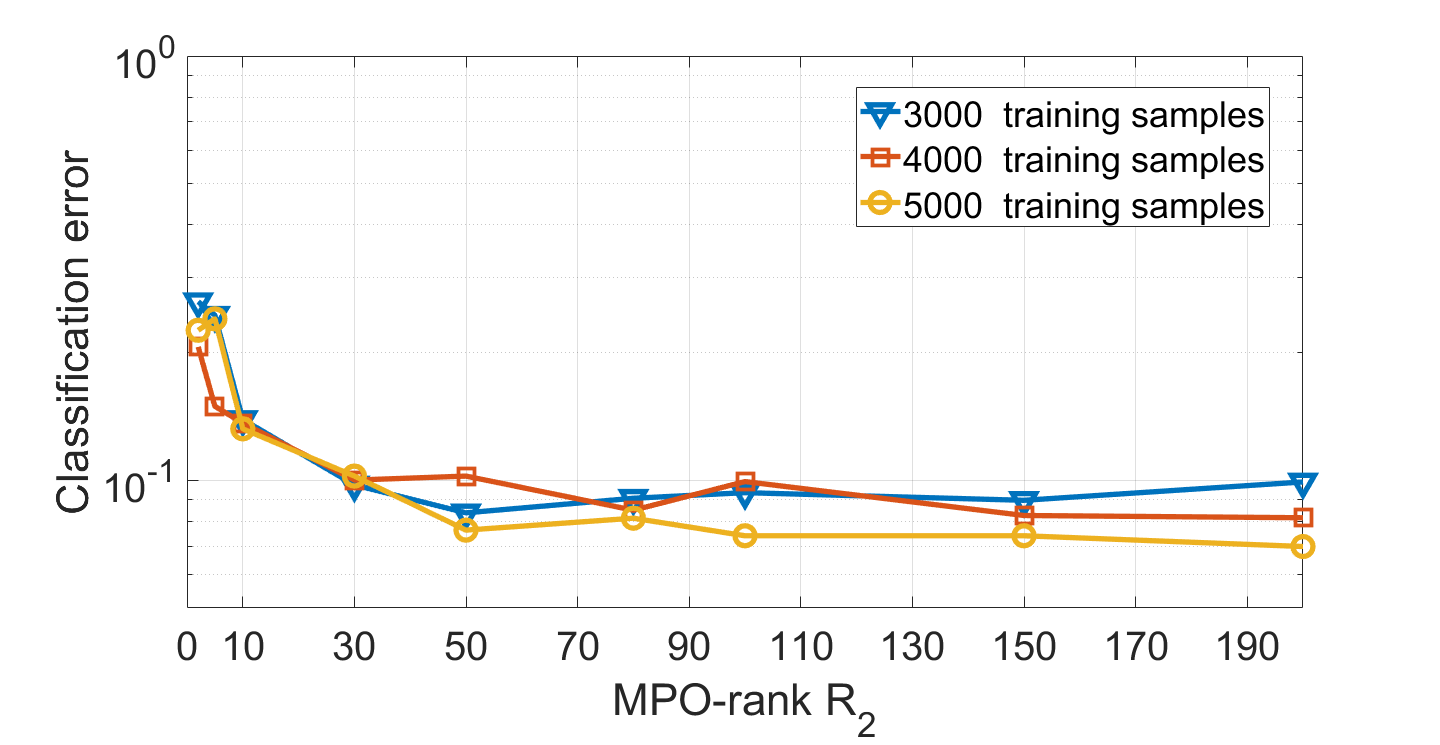}
\caption{Classification error as a function of the MPO-rank $R_2$ in the MPORBM model.}
\label{fig:experiment1.2}
\end{center}
\end{figure}

In this experiment, we demonstrate how the expressive power of the MPORBM is determined by the MPO-ranks. For this purpose, the MNIST dataset\footnote{http://yann.lecun.com/exdb/mnist/} was used to train an MPORBM for image classification. The MNIST dataset has a training set of $60000$ samples and a test set of $10000$ samples. Each sample is a $28\times 28$ grayscale picture of handwritten digits \{$0,\ldots,9$\}, where each pixel value was converted into a binary number. The hidden layer states in the RBM model were also regarded as features extracted from the training data and used in a $K$-NN classifier with $K=1$. The dimensions of the hidden layer were set to $M_1=M_2=10$. The MPO-rank $R_2$ was varied from 2 up to 200. To reveal the general rule of MPORBM model expression power on MPO-ranks, the above mentioned experiments were run for training sample sizes of 3000, 4000 and 5000, which are randomly chosen from training set. In addition, 10000 samples of the training set were chosen for validation. The classification error for each of these runs is shown in Fig.~\ref{fig:experiment1.2} as a function of $R_2$. Note that the MPORBM with $R_2=1$ is identical with the MvRBM for these matrix inputs. It can be seen that the classification error stabilizes when $R_2 \approx 40$, which indicates that a low-rank MPO may be powerful enough to model the latent probability distribution of the training data and a lot of storage and computational resources can be saved. We need to mention that by setting the MPO-ranks to their maximal values the MPORBM gets the same model expression ability as a standard RBM. This experiments, however, shows that using a low-rank MPORBM is more than enough to model real-life data.       

\subsection{Image completion and denoising}
\label{imageCom}

In this experiment, we show that an MPORBM is good at generative modeling by implementing image completion and denoising. We test the above generative tasks on the binarized MNIST dataset. Specifically, we randomly choose $50$ samples from the 60k training set to train the RBM models with $500$ iterations and using all $10000$ test samples to check the model generative performance. The standard RBM is set up with vectorial visual layer and hidden layer, while both MvRBM and MPORBM are constructed with 2-way tensorial visual layer and hidden layer. The MPO-rank in this experiment is set to 40 according to the observation from the previous experiment. The visual-hidden layer structure and total weight parameter number of each RBM model is listed in Table~\ref{tb:imageComParaNum}.

\begin{table}[h]
\footnotesize
\newcommand{\tabincell}[2]{\begin{tabular}{@{}#1@{}}#2\end{tabular}}
\centering
\caption{\label{tb:experiment3.1}The visual-hidden layer structure and total weight parameter number of different RBM models.}
\vspace{1ex}
\begin{tabular}{@{}lccc@{}}
  & RBM & MvRBM & MPORBM CD-AOP \\
\midrule
 Visual structure & $ 784 $ x $1$ &  $28$ x $28$ &  $28$ x $28$ \\
Hidden structure & $ 100 $ x $1$&  $10$ x $10$ &  $10$ x $10$\\
\# parameters & $ 78400 $&  $560$ &  $ 22400$ \\

 \end{tabular}
 \label{tb:imageComParaNum}
\end{table}

\begin{figure}[H]
\begin{center} 
\includegraphics[width=0.4\textwidth]{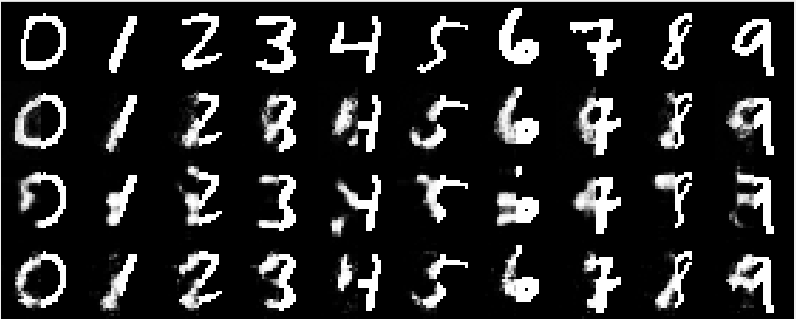}
\caption{Image completion result when given the right half image. First row: original binarized images; second row: RBM completed images; third row: MvRBM completed images; fourth row: MPORBM completed images. }
\label{fig:experiment3.11}
\end{center}
\end{figure}
\begin{figure}[H]
\begin{center} 
\includegraphics[width=0.4\textwidth]{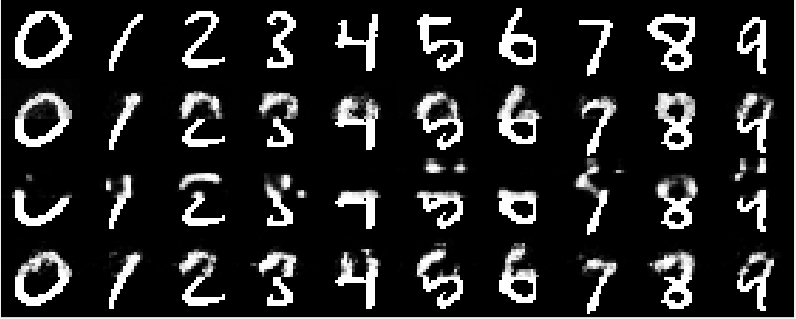}
\caption{Image completion result when given the bottom half image. First row: original binarized images; second row: RBM completed images; third row: MvRBM completed images; fourth row: MPORBM completed images. }
\label{fig:experiment3.12}
\end{center}
\end{figure}

In the image completion task, one half of the image is provided to the trained RBM models to complete the another half. As mentioned in~\cite{han2018unsupervised}, RBM completion performance is different when the given half image is in the column or row direction. Thus we investigate the completion ability of different RBM models with giving the right half and the bottom half image respectively. Figure~\ref{fig:experiment3.11} and Figure~\ref{fig:experiment3.12} demonstrate the completed images of different RBM models when given the same randomly selected right and bottom halfs, respectively. It is clear that MvRBM is not able to complete the image, which further confirms the necessity of the MPO generalization. Table~\ref{tb:imageCom} further lists the average PSNR results between the completed image and the original binarized image on the whole test set. We found that the MPORBM demonstrates a comparable image completion performance to a standard RBM, but with much fewer model parameter (around $29$\%).

\begin{table}[h]
\footnotesize
\newcommand{\tabincell}[2]{\begin{tabular}{@{}#1@{}}#2\end{tabular}}
\centering
\caption{\label{tb:experiment3.1}The average PSNR results on the whole test set when given the right and bottom half images respectively, and completing the left and upper half images.}
\vspace{1ex}
\begin{tabular}{@{}lccc@{}}
Given half   & RBM & MvRBM & MPORBM CD-AOP \\
\midrule
Right half & $ 13.86$~dB &  $12.35$~dB &  $ 13.79$~dB \\
Bottom half & $ 13.66$~dB &  $12.38$~dB & $ 13.69$~dB \\
 \end{tabular}
 \label{tb:imageCom}
\end{table}

In the image denoising task, we employ the same trained RBM models from the completion task and randomly add a certain percentage of salt \& pepper noise to the test set. The average PSNR results between the original binarized pictures and denoised pictures on the whole test set are listed in Table~\ref{tb:imageDenoise}. It is clear to see that the denoising performance of the MPORBM is comparable with a standard RBM when $10$\% noise is added, but performs better when higher percentages of noise are added. The fewer model parameters in the MPORBM may lead to a more robust generative model.    

\begin{table}[h]
\footnotesize
\newcommand{\tabincell}[2]{\begin{tabular}{@{}#1@{}}#2\end{tabular}}
\centering
\caption{\label{tb:experiment3.1}The average PSNR results on the whole test set when adding $p$\% salt \& pepper noises.}
\vspace{1ex}
\begin{tabular}{@{}lccc@{}}
$p$\% noise    & RBM & MvRBM & MPORBM CD-AOP \\
\midrule
$10$\% & $ 13.55$~dB &  $11.82$~dB & $13.49$~dB \\
$15$\% & $ 12.90$~dB &  $10.26$~dB & $13.24$~dB\\
$20$\% & $ 11.88$~dB &  $9.23$~dB & $12.95$~dB \\
 \end{tabular}
 \label{tb:imageDenoise}
\end{table}


\section{Conclusion}
\label{conclusion}
This paper has proposed the MPORBM, which preserves the tensorial nature of the input data and utilizes a matrix product operator (MPO) to represent the weight matrix. The MPORBM generalizes all existing RBM models to tensor inputs and has better storage complexity since the number of parameters grows only linearly with the order of the tensor. Furthermore, by representing both the visible and hidden layers as tensors, it is possible to retain useful structural information in the original data. In order to facilitate the exposition on the various computations, tensor network diagrams were introduced and used throughout the paper.

There are several avenues for future research. If both the visible and hidden layers are represented in a matrix product state (MPS) form, then all computations can be done on individual cores. This can significantly improve the computational complexity of the learning algorithm.

Furthermore, analogous to stacking RBMs into a Deep Belief Network (DBN) or Deep Belief Machine (DBM), MPORBMs can be stacked into an MPODBN or MPODBM to extend its expressive power and application. Again, this stacking can also gain computational benefits from keeping all layers in the MPS form.



\newpage
\small
\bibliographystyle{IEEEtran}
\bibliography{IEEEabrv,ijcai18}

\end{document}